# Research Progress of Convolutional Neural Network and its Application in Object Detection

Wei Zhang and Zuoxiang Zeng[*]

East China University of Science and Technology, 200237 Shanghai, China

*corresponding author：zengzx@ecust.edu.cn

**Abstract:** With the improvement of computer performance and the increase of data volume, the object detection based on convolutional neural network (CNN) has become the main algorithm for object detection. This paper summarizes the research progress of convolutional neural networks and their applications in object detection, and focuses on analyzing and discussing a specific idea and method of applying convolutional neural networks for object detection, pointing out the current deficiencies and future development direction.

## 1 Introduction

Object detection has a wide range of applications in areas such as intelligent video surveillance, automatic driving, and robot environment perception. Its main task is to locate the object of interest from the image and accurately determine the specific category and bounding box of each object. However, there are many factors that cause the object to deform in practice, which will make object detection very difficult, such as changes in posture and perspective, and occlusion of other objects. Traditional object detection methods are mainly divided into six steps: pre-processing, window sliding, feature extraction, feature selection, feature classification and post-processing. Among them, the most important thing is to improve the ability to express features and resist deformation, as well as the accuracy and speed of the classifier. Various forms of features have been proposed, such as SIFT(scale-invariant feature transform) [1]、Strip[2]，Haar[3] and HOG(histogram of oriented gradient) [4]；The main classifiers are: RF (random forest) [5], SVM(support vector machine) [6] and AdaBoost[7]。

However, these features are still low-level, with insufficient ability to express objects and poor separability, resulting in inefficient classifiers. Any single feature cannot be applied to multi-object detection. However, these features are still low-level, and the ability to express the object is insufficient and the separability is poor, resulting in inefficient classifiers. Any single feature cannot be applied to multi-object detection. For example, Haar can only be used for face detection, and HOG and Strip are used for pedestrian detection and vehicle detection respectively[8]. Therefore, Hinton [9] proposed the concept of deep learning in 2006, and a deep neural networks was used to automatically learn high-level features from a large amount of data according to a certain learning model, called learning features, and its expression ability is stronger. Commonly used deep learning models include auto encode (AE) [10], restricted Boltzmann machine (RBM) [11], and convolutional neural network ((convolutional neural network, CNN) [12]. Among them, the convolutional neural network integrates feature extraction, feature selection, and feature classification into the same model. Through end-to-end training, the function is optimized as a whole and the separability of features is enhanced. CNN-based object detection has become one of the research hotspots.

In this article, the basic ideas of convolutional neural networks will be briefly introduced, and the application research of CNN in object detection will be reviewed. On this basis, a specific idea and method of applying CNN for object detection will be analyzed and discussed, and its shortcomings and future development direction will be pointed out.

## 2 The basic structure and research progress of CNN

The convolutional neural network was first proposed by LeCun [13] as a classifier for image recognition. Its basic structure mainly includes an input layer, a convolutional layer, a pooling layer, a fully connected layer, and an output layer. The features of the convolutional layer are obtained from the local features of the previous layer through the weights shared by the convolution. In the convolutional neural network, the input image is subjected to feature extraction through multiple convolutional layers and pooling layers, and gradually changes from low-level features to high-level features. The high-level features are then classified by the fully connected layer and the output layer to generate a one-dimensional vector, which represents the current input image category. Therefore, according to the function of each layer, the convolutional neural network can be divided into two parts: feature extractor (including input layer, convolutional layer and pooling layer), classifier (including fully connected layer and output layer). The predecessors have done a lot of valuable work for these two parts through continuous optimization and update, which has continuously improved the efficiency of feature extraction and classification of convolutional neural networks.

Lin *et al.* [14] proposed the network structure of network in network, replacing the convolution kernel with a miniature multi-layer neural network, which can improve the ability of the network to fit complex nonlinear functions. Szegedy et al. [15] proposed the inception module, which sets convolution kernels of different sizes in the same convolutional layer to extract features with different scales on the previous layer of feature maps. Inspired by cell behavior, Simoncelli et al. [16] proposed a pooling layer model. When $P$=1, Lp pooling becomes mean pooling, and when $P=\infty$, it becomes maximum pooling. The research results show that Lp pooling provides better generalization capabilities than maximum pooling and mean pooling. In addition, Zeiler et al. [17] proposed stochastic pooling, which randomly selects the activation values of neurons in the local area according to the probability distribution to prevent overfitting of convolutional neural networks. Hinton et al. [18] proposed the use of the Dropout strategy in the fully connected layer. In the training phase, by randomly suppressing some neurons, the convolutional neural network only updates some of the neurons, thus the over-fitting problem of the fully connected layer can be effectively alleviated.

On the basis of the Dropout strategy, Wan et al. [19] further proposed the Drop Connect strategy, which randomly disconnected the neurons in the fully connected layer from the neurons in the previous layer during the training phase. So the generalization ability of convolutional neural network can be improved.

In addition to image detection and object classification, Ian GoodFellow [20] proposed a neural network-based generative model (Generative Adversarial Networks, GAN) in 2014. The model inverts the structure of a traditional neural network: its input is a set of low-dimensional noise, and the output is a synthetic image with spurious and real features. Unlike traditional neural networks, the GAN model implements reverse data generation from low to high dimensions. On this basis, Gao *et al.* [21] proposed an improved GAN model, which uses the Wasserstein distance instead of the traditional KL distance, and uses a game theory model for training, thereby improving the stability and convergence of GAN.

Convolutional neural networks are not only widely used in object recognition and object classification, but also play a key role in data generation and data transformation, such as style transfer of paintings [22], emotional transformation of human speech [23] and image super-resolution reconstruction [24].

## 3 Research progress of object detection based on CNN

As early as 1994, the convolutional neural network was successfully applied to object detection [25], but it did not become the mainstream. In 2012, the convolutional neural network AlexNet [26] made a major breakthrough in image recognition, laying its core position in object detection.

Since the convolutional neural network itself has the functions of feature extraction, feature selection and feature classification, the candidate area generated by each sliding window can be directly classified by the convolutional neural network to determine whether it is an object to be detected. Therefore, the object detection based on convolutional neural network contains only three steps: window sliding, image classification and post-processing. The key to improving the accuracy of image recognition is how to improve the feature extraction ability, feature selection ability, and feature classification ability of convolutional neural networks.

In 1998, LeCun et al. [13] proposed LeNet-5, which was successfully used to recognize handwritten digital images. As an early convolutional neural network, LeNet-5 contains fewer layers, with only 2 convolutional layers, 2 pooling layers, and 3 fully connected layers. The model input is a single-channel 3232-sized image, and the output is a l0-dimensional vector. LeNet-5 has about 60,000 training parameters. The lack of data and overfitting at that time limited the application of the model in other fields. Krizhevsky et al.[27] proposed AlexNet, which greatly improved the accuracy of classification on the Image Net dataset. Compared with LeNet 5, the neural network, which has 60 million parameters and 650,000 neurons, consists of five convolutional layers, some of which are followed by max-pooling layers, and three fully connected layers with a final 1000-way softmax. To make training faster, the nonsaturating neurons and a very efficient GPU implementation of the convolution operation were used. To reduce overfitting in the fully connected layers, a recently developed regularization method called "dropout" were employed.

To solve the problem of long training time and high resource share of deep convolutional neural network, Simonyan et al. [28] designed a shallow convolutional neural network to achieve the similar classification accuracy. The deep and shallow CNNs have data pre-processing, feature extraction and softmax classification. Moreover, the proposed convolutional neural networks can be integrated and implemented in other colour image database.

Szegedy etal.[15] proposed a deep convolutional neural network architecture codenamed "Inception", which was responsible for setting the new state of the art for classification and detection in ILSVRC 2014. One particular incarnation is called GoogLeNet which contains 22 convolutional layers and 5 pooling layers. The multiple inception modules was built to increase the depth and width of the CNN. The main hallmark of this architecture is the improved utilization of the computing resources inside the network. Due to the small size of the convolution kernel in the inception module, the training parameters of Goog LeNet are only 1/12 of the number of AlexNet parameters, but the accuracy of image recognition on the Image Net dataset is increased by about 10%.

He *et al.* [29] proposed a deep residual learning framework to solve the degradation problem by using a multi-layer network to fit a residual mapping. In the convolutional neural network, the low-level feature map is directly mapped to the high-level feature map, and the identity mapping is established to make the convolutional layer perform residual learning, which solves the degradation problem of the deep convolutional neural network to a certain extent.

Vaillant *et al.* [25] first proposed the application of convolutional neural network to face detection. Garcia *et al.* [30] added a convolutional layer and a pooling layer to improve the expression ability of features, and adopted a training strategy based on Boosting to accelerate the convergence speed of convolutional neural networks and enhance face detection Robustness of the device.

Sermanet *et al.* [31] proposed the application of convolutional neural networks to pedestrian detection. It is characterized by first pre-training each convolutional layer unsupervised by convolutional sparse coding, and then using training samples to supervisely adjust a convolutional neural network, which significantly improves the ability of features to express the object.

Ouyang *et al.* [32] proposed a joint pedestrian detection model based on convolutional neural networks. The model integrates feature extraction, deformation processing, occlusion processing, and feature classification into a single convolutional neural network. The relationship between pedestrian and components can be automatically established through end-to-end training, and the separability of features is enhanced. Chen et al. [33] proposed the concept of face candidate area. Adboost face detector is used to predict all candidate areas, and the candidate areas that may be faces are reserved; then a small-scale convolutional neural network is used to judge Whether the candidate area is a human face, and a medium-scale convolutional neural network is used to complete the classification of all candidate areas.

Oquab *et al.* [34] proposed a method of transferring mid-level features of CNNs, which applied features used for image classification to object detection. The highlight of this method is to add two adaptive layers after the AlexNet fully connected layer, and use a small amount of training samples to adjust the parameters of these two adaptive layers to train for object detection.

Because the expression ability and separability of the learned features used in the above methods are better than the designed features, so the object detection based on the CNN has achieved a higher accuracy than the traditional object detection methods. However, since the object may be located at any position of the image to be detected, and the size of the object is uncertain, it is usually necessary to construct an image pyramid of the image to be detected, slide the window on multiple scales, and search for the object position in an exhaustive manner. As a result, the number of candidate regions is huge and the detection speed is very slow.

In order to reduce the number of candidate areas, researchers found that specific algorithms can be used to extract sub-images with a certain semantic meaning from the images to be detected as candidate areas. The number of these candidate regions is small and the size is not fixed. After the classification and recognition of the convolutional neural network, multi-scale and multi-category object detection is realized, which greatly improves the efficiency of object detection. Commonly used candidate area extraction methods include ICOP [35], CPMC [36], selective search [37] and so on.

Gao et al. [38] proposed an image matching method with rotation and scale invariance. The model first established sparse key point matching based on local invariance features, and then used the sparse matching result as a reference to complete dense matching. Finally, the dense matching relationship of the wide baseline image is obtained. According to the matching relationship, the depth information in the two-dimensional image can be recovered, so as to realize the three-dimensional reconstruction of the object scene. Shi et al. [39] made the method affine invariance by extracting the affine invariant features in the image, and improved the robustness of the model. Girshick et al. [40] proposed the R-CNN model. The model uses the selective search method to extract several candidate regions from the image to be detected; then converts the candidate regions into a uniform size, uses convolutional neural networks to extract features, and finally uses multiple SVMs to classify features to complete multiple object detection.

Due to the selection of a better candidate area and the use of AlexNet to determine the category of the candidate area, the recognition effects of the above methods are very good, which greatly improves the detection effect on the Pascal VOC database.

In order to enable the CNN to process candidate regions of arbitrary size, He et al. [41] designed a spatial pyramid pooling layer in the convolutional neural network. This layer can extract feature vectors of the same length from feature maps of different sizes, which overcomes the limitation that convolutional neural networks can only accept fixed-size inputs.

Girshick [42] proposed the fast R-CNN model in order to improve the detection speed and accuracy of the R-CNN model. This model still uses the selective search method to extract several candidate regions from the image to be detected. Compared with the R-CNN model extracting features for each candidate area, fast R-CNN only extracts features from the image to be detected; then the feature map corresponding to the candidate area is mapped to a fixed-length feature vector through the spatial pyramid pooling layer; and finally the features are classified by a fully connected neural network, and the coordinates of the bounding box are predicted to correct the candidate area.

Ren et al. [43] proposed a faster R-CN model to speed up the calculation speed of extracting candidate regions. The model consists of a convolutional neural network (region proposal network, RPN) for extracting candidate regions and a convolutional neural network fast R-CNN for object detection. RPN predicts whether there is an object at each position in the image to be detected, and gives the most likely candidate area. Fast R-CNN discriminates the categories of all candidate regions and predicts the bounding box of the object to correct the candidate regions. Because RPN and fast R-CNN share the feature extraction part of the convolutional neural network, the model only needs to extract the features of the image to be detected once, which speeds up the object detection.

These methods rely heavily on the accuracy of candidate area extraction. If the detection scene is complex and the object is not obvious, it is difficult for the candidate area extraction method to capture the area near the object, resulting in the object not being detected.

## 4 Conclusion

This paper analyzes the problems of traditional object detection methods, the basic structure, research status and common models of convolutional neural networks. The main research methods and research progress of object detection based on convolutional neural network are summarized, and the existing problems are analyzed. In short, object detection based on convolutional neural networks is a challenging subject, which has very important research significance and application value. Although research in this area has achieved some success, it is still far from practical application. Nowadays, the expansion of data volume and the improvement of hardware performance provide new opportunities and challenges for this subject, so it is worthy of more in-depth research.